\documentclass[11pt,a4paper]{article}
\usepackage[hyperref]{acl2021}
\usepackage{times}
\usepackage{latexsym}

\usepackage{url}
\usepackage[title]{appendix}
\usepackage{footnote}

\usepackage{dblfloatfix}
\usepackage{amssymb}
\usepackage{graphicx}
\usepackage{xcolor,pict2e}
\usepackage{float}
\usepackage{amsmath}
\usepackage{hyperref}
\usepackage{xcolor}
\definecolor{ForestGreen}{cmyk}{1,0.1,1,0}
\usepackage{subcaption}
\usepackage{multirow}
\usepackage{mathabx}
\usepackage{bm}

\usepackage{microtype}

\aclfinalcopy 

\title{Improving the Faithfulness of Attention-based Explanations with Task-specific Information for Text Classification}

\author{George Chrysostomou \; \;\; \; \; \;\; \; Nikolaos Aletras\\
  Department of Computer Science, University of Sheffield \\
  United Kingdom \\
  \texttt{\{gchrysostomou1, n.aletras\}@sheffield.ac.uk} \\}

\date{}

\begin{document}
\maketitle

\begin{abstract}

Neural network architectures in natural language processing often use attention mechanisms to produce probability distributions over input token representations. Attention has empirically been demonstrated to improve performance in various tasks, while its weights have been extensively used as explanations for model predictions. Recent studies \citep{jain2019attention,serrano2019attention,wiegreffe2019attention} have showed that it cannot generally be considered as a faithful explanation \citep{jacovi-goldberg-2020-towards} across encoders and tasks. In this paper, we seek to improve the faithfulness of attention-based explanations for text classification. We achieve this by proposing a new family of Task-Scaling (TaSc) mechanisms that learn task-specific non-contextualised information to scale the original attention weights. Evaluation tests for explanation faithfulness, show that the three proposed variants of TaSc improve attention-based explanations across two attention mechanisms, five encoders and five text classification datasets without sacrificing predictive performance. Finally, we demonstrate that TaSc consistently provides more faithful attention-based explanations compared to three widely-used interpretability techniques.\footnote{Code for all experiments will be publicly released.}
\end{abstract}

\section{Introduction}

Natural Language Processing (NLP) approaches for text classification are often underpinned by large neural network models~\citep{cho-etal-2014-learning,Devlin2019}. Despite the high accuracy and efficiency of these models in dealing with large amounts of data, an important problem is their increased complexity that makes them opaque and hard to interpret by humans which usually treat them as \emph{black boxes} \citep{zhang2018opening,blackboxnlp}.

Attention mechanisms~\citep{bahdanau2014neural} produce a probability distribution over the input to compute a vector representation of the entire token sequence as the weighted sum of its constituent vectors. A common practice is to provide explanations for a given prediction and qualitative model analysis by assigning importance to input tokens using scores provided by attention mechanisms~\citep{chen-etal-2017-recurrent-attention,wang-etal-2016-attention,jain-etal-2020-learning,sun-lu-2020-understanding} as a mean towards model interpretability~\citep{lipton2016mythos,Miller2017}. 


A faithful explanation is one that accurately represents the true reasoning behind a model's prediction~\citep{jacovi-goldberg-2020-towards}. A series of recent studies illustrate that explanations obtained by attention weights do not always provide faithful explanations~\citep{serrano2019attention} while different text encoders can affect attention interpretability, e.g. results can differ when using a recurrent or non-recurrent encoder \citep{wiegreffe2019attention}.  


A limitation of attention as an indicator of input importance is that it refers to the word in context due to information mixing in the model \citep{tutek-snajder-2020-staying}. Motivated by this, we aim to improve the effectiveness of neural models in providing more \emph{faithful} attention-based explanations for text classification, by introducing non-contextualised information in the model. Our contributions are as follows:

\begin{itemize}
    \item We introduce three Task-Scaling (TaSc) mechanisms ($\S$\ref{sec:TaSc}), a family of encoder-independent components that learn task-specific non-contextualised importance scores for each word in the vocabulary to scale the original attention weights which can be easily ported to any neural architecture;
    \item We show that TaSc variants offer more robust, consistent and faithful attention-based explanations compared to using vanilla attention in a set of standard interpretability benchmarks, without sacrificing predictive performance ($\S$\ref{sec:results});
    \item We demonstrate that attention-based explanations with TaSc consistently outperform explanations obtained from two gradient-based and a word-erasure explanation approaches ($\S$\ref{sec:results_tasc_vs_interpretability}).
     

\end{itemize}




\section{Related Work}
\label{sec:related_work}

\subsection{Model Interpretability}



Explanations for neural networks can be obtained by identifying which parts of the input are important for a given prediction. One way is to use sparse linear meta-models that are easier to interpret~\citep{ribeiro2016model,Lundberg2017,nguyen2018comparing}. Another way is to calculate the difference in a model's prediction between keeping and omitting an input token  \citep{robnik2008explaining, li2016understanding, nguyen2018comparing}. Input importance is also measured using the gradients computed with respect to the input~\citep{kindermans2016investigating,li-etal-2016-visualizing,larras2016explaining, integrated_gradients}. \citet{chen-ji-2020-learning} propose learning a variational word mask to improve model interpretability. Finally, extracting a short snippet from the original input text (rationale) and using it to make a prediction has been recently proposed \citep{lei-etal-2016-rationalizing, Bastings2019,treviso-martins-2020-explanation, jain-etal-2020-learning, chalkidis2021paragraph}. 

\citet{nguyen2018comparing} and \citet{atanasova2020diagnostic} compare explanations produced by different approaches, showing that in most cases gradient-based approaches outperform sparse linear meta-models. 

\subsection{Attention as Explanation}


Attention weights have been extensively used to interpret model predictions in NLP; i.e. \citep{cho-etal-2014-learning, pmlr-v37-xuc15,barbieri-etal-2018-interpretable, ghaeini2018interpreting}. 
However, the hypothesis that attention should be used as explanation had not been explicitly studied until recently.

\citet{jain2019attention} first explored the effectiveness of attention explanations. They show that adversary attention distributions can yield equivalent predictions with the original attention distribution, suggesting that attention weights do not offer robust explanations. 
 In contrast to \citet{jain2019attention}, \citet{wiegreffe2019attention} and \citet{Vashishth2019} demonstrate that attention weights can in certain cases provide robust explanations. \citet{pruthi-etal-2020-learning} also investigate the ability of attention weights to provide plausible explanations. They test this through manipulating the attention mechanism by penalising words a priori known to be relevant to the task, showing that the predictive performance remain relatively unaffected. \citet{sen-etal-2020-human} assess the plausibility of attention weights by correlating them with manually annotated explanation heat-maps, where plausibility refers to how convincing an explanation is to humans \citep{jacovi-goldberg-2020-towards}. However, \citet{jacovi-goldberg-2020-towards} and \citet{grimsley-etal-2020-attention} suggest caution with interpreting the results of these experiments as they do not test the faithfulness of explanations (e.g. an explanation can be non-plausible but faithful or vice-versa).


\citet{serrano2019attention} test the faithfulness of attention-based explanations by removing tokens to observe how fast a {\it decision flip} happens. Results show that gradient attention-based rankings (i.e. combining an attention weight with its gradient) better predict word importance for model predictions, compared to just using the attention weights. \citet{tutek-snajder-2020-staying} propose a method to improve the faithfulness of attention explanations when using recurrent encoders by introducing a word-level objective to sequence classification tasks. Focusing also on recurrent-encoders, \citet{mohankumar-etal-2020-towards} introduce a modification to recurrent encoders to reduce repetitive information across different words in the input to improve faithfulness of explanations. 

To the best of our knowledge, no previous work has attempted to improve the faithfulness of attention-based explanations across different encoders for text classification by inducing task-specific information to the attention weights.



\section{Neural Text Classification Models}

In a typical neural model with attention for text classification; one-hot-encoded tokens $x_i \in \rm \!R^{\vert V \vert}$ are first mapped to embeddings $\mathbf{e}_i \in \rm \!R^{d}$, where $i \in [1,...,t]$ denotes the position in the sequence, $t$ the sequence length, $\vert V \vert$ the vocabulary size and $d$ the dimensionality of the embeddings. The embeddings $\mathbf{e}_i$ are then passed to an encoder to produce hidden representations $ \mathbf{h}_i = Enc(\mathbf{e}_i)$, where $\mathbf{h}_i \in \rm \!R^{N}$, with $N$ the size of the hidden representation. A vector representation $\mathbf{c}$ for the entire text sequence {$x_1,...,x_t$} is subsequently obtained as the sum of $\mathbf{h}_i$ weighted by attention scores $\alpha_i$: 
\begin{align}
    \mathbf{c} = \sum_i \mathbf{c}_i, \quad \mathbf{c}_i = \mathbf{h}_i\alpha_i, \quad \mathbf{c} \in \rm \!R^{N} 
    \label{eq:context_vector_1}
\end{align}
Vector $\mathbf{c}$ is finally passed to the output, a fully-connected linear layer followed by a softmax activation function. 

\subsection{Encoders}

To obtain representations $\mathbf{h}_i$, we consider the following recurrent, non-recurrent and Transformer~\cite{vaswani2017attention} encoders,
$Enc(.)$,
as in \citep{jain2019attention,wiegreffe2019attention}: (i) bidirectional Long Short-Term Memory ({\bf LSTM}; \citet{lstm}); (ii) bidirectional Gated Recurrent Unit ({\bf GRU}; \citet{cho-etal-2014-learning}); (iii) Convolutional Neural Network ({\bf CNN}; \citet{lecun1999object}); (iv) Multi-Layer Perceptron ({\bf MLP}); (v) \textbf{BERT}\footnote{We use BERT to obtain $\mathbf{h}_i$ with an attention mechanism on top for consistency with the other encoders} \citep{Devlin2019}.

\subsection{Attention Mechanisms}
\label{sec:attention}

Attention scores ($a_i$) are computed by passing the representations ($\mathbf{h}_i$) obtained from the encoder to the attention mechanism which usually consists of a similarity function $\phi$ followed by softmax:
\begin{align}
    a_{i} = \frac{\text{exp}(\phi(\mathbf{h_i},\mathbf{q}))}{\sum_{k=1}^t \text{exp}(\phi(\mathbf{q},\mathbf{h_k}))}
\end{align}

\noindent where $\mathbf{q} \in \!R^{N}$ is a trainable self-attention vector similar to \citet{Yang2016}.  

Following \citet{jain2019attention}, we consider two self-attention similarity functions: (i) \textbf{Additive Attention (Tanh; }\citet{bahdanau2014neural}\textbf{):} 
\begin{align}
     \phi(h_i, \mathbf{q}) = \mathbf{q}^T \tanh (W\mathbf{h}_i)
\end{align}
\noindent where $W$ is a trainable model parameter; and (ii) \textbf{Scaled Dot-Product (Dot; }\citet{vaswani2017attention}\textbf{):} 
\begin{align}
     \phi(h_i, \mathbf{q}) = \frac{\mathbf{h}_i^T\mathbf{q}}{\sqrt{N}} 
\end{align}

\section{Task-Scaling (TaSc) Mechanisms \label{sec:TaSc}}



Attention indicates how well inputs around a position $i$ correspond to the output \citep{bahdanau2014neural}. 
For example, in a bidirectional recurrent encoder each token representation $\mathbf{h}_i$ contains information from the whole sequence so the attention weights actually refer to the input word \emph{in context} and not individually~\citep{tutek-snajder-2020-staying}.



Inspired by the simple and highly interpretable bag-of-words models, which assign a single weight for each word type (word in a vocabulary), we hypothesise that by scaling each input word's contextualised representation  $\mathbf{c}_{i}$ (see Eq. \ref{eq:context_vector_1}) by its attention score \textit{and} and a non-contextualised word type scalar score, we can improve
attention-based explanations. The intuition is that by having a less contextualised sequence representation $\mathbf{c}$ we can reduce information mixing for attention. 

For that purpose, we introduce the non-contextualised word type score $s_{x_i}$ in Eq.~\ref{eq:context_vector_1} to enrich the text representation $\mathbf{c}$, such that:
\begin{equation}
    \mathbf{c} = \sum_i \mathbf{h}_i\alpha_i s_{x_i} , \quad \mathbf{c}  \in \rm \!R^{N}
\end{equation}
We compute $s_{x_i}$ by proposing three Task-Scaling ({\bf TaSc}) mechanisms.\footnote{Number of parameters for each proposed mechanism in Appendix \ref{sec:additional_parameters}.} 


\subsection{Linear TaSc (Lin-TaSc)}


We first introduce Linear TaSc (Lin-TaSc), the simplest method in the family of TaSc mechanisms that estimates a scalar weight for each word in the vocabulary by introducing a new vector $\mathbf{u} \in \rm \!R^{\vert V \vert}$. Given the input sequence $\mathbf{x} = [x_1, \hdots, x_t]$ representing one-hot-encodings of the tokens, we perform a look up on $\mathbf{u}$ to obtain the scalar weights of words in the sequence. $\mathbf{u}$ is randomly initialised and updated partially at each training iteration, because naturally each input sequence contains only a small subset of the vocabulary words.


We then obtain a task-scaled embedding $\mathbf{\hat{e}}_i$ for a token $i$ in the input by multiplying the original token embedding with its word type weight $u_i$:
\begin{equation}
  \mathbf{\hat{e}}_i = u_i\mathbf{e}_i  
  \label{eq:lin_tasc}
\end{equation}

The intuition is that the embedding vector $\mathbf{e}_i$ was trained on general corpora and is a non-contextualised ``generic'' representation of input $x_i$. As such the score $u_i$ will scale $\mathbf{e}_i$ to the task. We subsequently compute context-independent scores $s_{x_i}$ for each token in the sequence, by summing all elements of its corresponding task-scaled embedding $\mathbf{\hat{e}}_i$; $s_{x_i} = \sum^d \mathbf{\hat{e}}_i$ in a similar way that token embeddings are averaged in the top-layers of a neural architecture. We opted to sum-up and not average, because we want to retain large and small values from the task-scaled embedding vector $\mathbf{\hat{e}}_i$ \citep{atanasova2020diagnostic}.\footnote{We also tried max and mean-pooling or using the $u_i$ directly instead of $s_i$ in early experimentation resulting in lower results. \label{max-pool-footnote}}


As the attention scores pertain to the word in context \citep{tutek-snajder-2020-staying}, we also expect the score $s_{x_i}$ to pertain to the word without the contextualised information. That way, we complement attention which results into a richer sequence representation $\mathbf{c}$.

\subsection{Feature-wise TaSc (Feat-TaSc)}

Lin-TaSc assigns equal weighting to all the dimensions of the word embedding $\mathbf{e}_i$ (see Eq.~\ref{eq:lin_tasc}), but some of them might be more important than others. Inspired by the RETAIN mechanism \citep{NIPS2016_6321}, Feature-wise TaSc (Feat-TaSc) learns different weights for each embedding dimension to identify the most important of them. Compared to Lin-TaSc where $\mathbf{e}_i$ is scaled uniformly across all vector dimensions, with Feat-TaSc each dimension is scaled independently. To achieve this, we introduce a learnable matrix $\mathbf{U} \in \rm \!R^{\vert V \vert \times d}$. Similar to Lin-TaSc, given the input sequence $\mathbf{x}$, we perform a look up on $\mathbf{U}$ to obtain $\mathbf{U}_s = [\mathbf{u}_1, \hdots, \mathbf{u}_t]$. $\mathbf{U}$ is randomly initialised and updated partially at each training iteration. 
To obtain $s_{x_i}$, we perform a dot product between $\mathbf{u}_i$ and embedding vector $\mathbf{e}_i$; $s_{x_i} = \mathbf{u}_i \cdot \mathbf{e}_i$.

\subsection{Convolutional TaSc (Conv-TaSc)}

Lin-TaSc and Feat-TaSc weigh the original word embedding $\mathbf{e}_i$ but do not consider any interactions between embedding dimensions. Conv-TaSc addresses this limitation by extending Lin-TaSc.\footnote{We only apply Conv-TaSc over Lin-TaSc to keep the mechanism relatively lightweight. Note that Feat-TaSc learns an extra matrix of equal size to the embedding matrix.} We apply a CNN\footnote{See CNN configurations in Appendix \ref{sec:appendix_model_hyperparameters}.} with $n$ channels over the scaled embedding $\mathbf{\hat{e}}_i$ from Lin-TaSc, keeping a single stride and a 1-dimensional kernel. This way, we ensure that input words remain context-independent. We then sum over the filtered scaled embedding $\mathbf{\hat{e}}_i^f$, to obtain the scores $s_{x_i}$; $s_{x_i} = \sum^d \mathbf{\hat{e}}_i^f$.\textsuperscript{\ref{max-pool-footnote}}

\section{Evaluating Attention-based Interpretability}
\label{sec:interpretability}


\citet{jacovi-goldberg-2020-towards} propose that an appropriate measure of \emph{faithfulness} of an explanation can be obtained through \emph{erasure} (the most relevant parts of the input--according to the explanation--are removed). We therefore follow this evaluation approach similar to \citet{serrano2019attention}, \citet{atanasova2020diagnostic} and \citet{nguyen2018comparing}.\footnote{Note that \citet{jacovi-goldberg-2020-towards} argue that a human evaluation is not an appropriate method to test faithfulness.}

\subsection{Attention-based Importance Metrics} 

We opt using the following three input importance metrics by \citet{serrano2019attention}:\footnote{\citet{serrano2019attention} show that gradient-based attention ranking metrics ($\nabla \bm{\alpha}$, $\bm{\alpha} \nabla \bm{\alpha}$) are better in providing faithful explanations compared to just using attention ($\bm{\alpha}$).}

\begin{itemize}
    \item $\bm{\alpha}$:  Importance rank corresponding to normalised attention scores. 

    \item $\nabla \bm{\alpha}$: Provides a ranking by computing the gradient of the predicted label $\hat{y}$ with respect to each attention score $\alpha_i$ in descending order, such that $\nabla \alpha_i = \frac{\partial \hat{y}}{\partial \alpha_i}$.

    \item $\bm{\alpha} \nabla \bm{\alpha}$: Scales the attention scores $\alpha_i$ with their corresponding gradients $\nabla \alpha_i$.
    
\end{itemize}

\subsection{Faithfulness Metrics} 


\paragraph{Decision Flip - Most Informative Token:} The average percentage of decision flips (i.e. changes in model prediction) occurred in the test set by removing the token with highest importance.
    
\paragraph{Decision Flip - Fraction of Tokens:} The average fraction of tokens required to be removed to cause a decision flip in the test set.

Note that we conduct all experiments at the input level (i.e. by removing the token from the input sequence instead of only removing its corresponding attention weight) as we consider the scores from importance metrics to pertain to the corresponding input token following related work \citep{larras2016explaining,arras2017explaining,nguyen2018comparing, Vashishth2019,grimsley-etal-2020-attention, atanasova2020diagnostic}.

\section{Experiments and Results}
\label{sec:results}

\subsection{Data}

\renewcommand*{\arraystretch}{1.0}
\begin{table}[!ht]
\small
\centering
\begin{tabular}{l|cc|c}
\textbf{Dataset}   & Av. $|W|$ & $\mathbf{|V|}$ & \textbf{\begin{tabular}[c]{@{}c@{}}Splits\\   Train/Dev/Test\end{tabular}}  \\ \hline
SST                & 20                  & 13,686     & 6,920 / 872 / 1,821                                                                                                                                \\
ADR         & 22                  & 6,716    & 14,452 / 2,551 / 4,251                                                                                                                              \\
IMDB               & 185                 & 12,147     & 17,212 / 4,304 / 4,363                                                                                                                                \\
AG             & 34                  & 14,573     & 60,895 / 7,145 / 3,960                                                                                                                              \\
MIMIC  & 2,180                & 16,277     & 4,654 / 822 / 1,369                                                                                                                                  \\
                     
\end{tabular}
\caption{Dataset statistics including average words per instance, vocabulary size and splits.}
\label{data_characteristics}
\end{table}

We use five datasets for text classification following~\citet{jain2019attention}: (i) \textbf{SST} \citep{socher2013sst}; (ii) \textbf{IMDB} \citep{maas2011learning}; (iii) \textbf{ADR} Tweets \citep{sarker2015utilizing}; (iv) \textbf{AG} News;\footnote{\url{https://di.unipi.it/~gulli/AG_corpus_of_news_articles.html}} and (v) \textbf{MIMIC} Anemia \citep{johnson2016mimic}. See Table \ref{data_characteristics} for detailed data statistics.

\subsection{Predictive Performance}

A prerequisite of interpretability is to obtain robust explanations without sacrificing predictive performance \citep{lipton2016mythos}. Table \ref{tab:data_performances} shows the macro F1-scores of all models across datasets, encoders and attention mechanisms using the three TaSc variants (Lin-TaSc, Feat-TaSc and Conv-TaSc described in Section~\ref{sec:TaSc}) and without TaSc (No-TaSc).\footnote{For model hyper-parameters and prepossessing steps see Appendix \ref{sec:appendix_model_hyperparameters}.} 

In general, all TaSc models obtain comparable performance and in some cases outperform No-TaSc across datasets and attention mechanisms. However, our main aim is not to improve predictive performance but the faithfulness of attention-based explanations, which we illustrate below.

\begin{table}[!ht]
\setlength\tabcolsep{1pt}
\small
\centering
\begin{tabular}{cc|cc|cc|cc|cc}
\multicolumn{1}{c}{\textbf{Data}}              & \multicolumn{1}{c|}{\textbf{Enc()}}             & \multicolumn{2}{c|}{\textbf{No-TaSc}}                         & \multicolumn{2}{c|}{\textbf{Lin-TaSc}}                                       & \multicolumn{2}{c|}{\textbf{Feat-TaSc}}                                                                          & \multicolumn{2}{c}{\textbf{Conv-TaSc}}                                                                         \\
\textbf{}& \textbf{} & \textbf{Dot} & \textbf{Tanh} & \textbf{Dot} & \textbf{Tanh}   &  \textbf{Dot} & \textbf{Tanh} &  \textbf{Dot} & \textbf{Tanh} \\ \hline
\multirow{5}{*}{SST}              &  BERT & .91 & .90 &              .89 &              .88 &              .85 &              .88 &  \underline{.91} &     \textbf{.91} \\
          &    LSTM & .76 & .75 &     \textbf{.79} &     \textbf{.79} &     \textbf{.79} &      \textbf{.80} &     \textbf{.78} &     \textbf{.77} \\
          &     GRU & .76 & .77 &     \textbf{.79} &     \textbf{.78} &      \textbf{.80} &     \textbf{.79} &     \textbf{.77} &  \underline{.77} \\
          &     MLP & .76 & .76 &     \textbf{.78} &     \textbf{.78} &     \textbf{.79} &     \textbf{.78} &     \textbf{.79} &     \textbf{.79} \\
         &     CNN & .76 & .74 &      \textbf{.80} &     \textbf{.78} &      \textbf{.80} &      \textbf{.80} &     \textbf{.78} &     \textbf{.76} \\ \hline
         
\multirow{5}{*}{ADR}              & BERT & .80 & .79 &              .78 &              .77 &              .79 &              .76 &              .78 &              .77 \\
      &    LSTM & .74 & .73 &     \textbf{.75} &     \textbf{.75} &  \underline{.74} &     \textbf{.75} &              .73 &     \textbf{.75} \\
      &     GRU & .74 & .73 &     \textbf{.76} &     \textbf{.75} &  \underline{.74} &     \textbf{.76} &  \underline{.74} &     \textbf{.75} \\
      &     MLP & .74 & .68 &     \textbf{.75} &     \textbf{.74} &     \textbf{.75} &     \textbf{.74} &     \textbf{.75} &     \textbf{.74} \\
      &     CNN & .73 & .69 &     \textbf{.75} &     \textbf{.74} &     \textbf{.74} &     \textbf{.75} &     \textbf{.76} &     \textbf{.75} \\ \hline
      
\multirow{5}{*}{IMDB}             & BERT & .93 & .93 &  \underline{.93} &              .92 &              .92 &              .92 &  \underline{.93} &  \underline{.93} \\
         &    LSTM & .89 & .89 &              .88 &              .88 &              .88 &  \underline{.89} &  \underline{.89} &  \underline{.89} \\
         &     GRU & .89 & .90 &              .88 &              .88 &  \underline{.89} &              .89 &  \underline{.89} &              .89 \\
         &     MLP & .88 & .88 &  \underline{.88} &  \underline{.88} &  \underline{.88} &  \underline{.88} &     \textbf{.89} &  \underline{.88} \\
         &     CNN & .88 & .88 &  \underline{.88} &  \underline{.88} &  \underline{.88} &  \underline{.88} &  \underline{.88} &     \textbf{.89} \\ \hline
        
\multirow{5}{*}{AG}               & BERT & .94 & .94 &  \underline{.94} &  \underline{.94} &  \underline{.94} &  \underline{.94} &  \underline{.94} &  \underline{.94} \\
       &    LSTM & .92 & .93 &  \underline{.92} &              .92 &  \underline{.92} &              .92 &  \underline{.92} &              .92 \\
       &     GRU & .92 & .92 &  \underline{.92} &  \underline{.92} &  \underline{.92} &  \underline{.92} &  \underline{.92} &  \underline{.92} \\
       &     MLP & .92 & .92 &  \underline{.92} &  \underline{.92} &              .91 &              .91 &  \underline{.92} &  \underline{.92} \\
       &     CNN & .92 & .92 &  \underline{.92} &  \underline{.92} &  \underline{.92} &  \underline{.92} &  \underline{.92} &  \underline{.92} \\\hline
      
\multirow{5}{*}{MIMIC}            & BERT\footnotemark  & .82 & .84 &  \underline{.82} &              .83 &     \textbf{.83} &              .83 &     \textbf{.83} &              .83 \\
  &    LSTM & .87 & .89 &  \underline{.87} &              .87 &     \textbf{.88} &              .88 &     \textbf{.88} &              .88 \\
  &     GRU & .87 & .89 &  \underline{.87} &              .88 &     \textbf{.88} &              .88 &     \textbf{.88} &              .88 \\
  &     MLP & .87 & .87 &  \underline{.87} &              .86 &              .86 &              .86 &  \underline{.87} &              .86 \\
  &     CNN & .88 & .89 &  \underline{.88} &              .87 &              .87 &              .87 &  \underline{.88} &              .88                                              
\end{tabular}
\caption{F1-macro average scores (3 runs) across datasets, encoders and attention mechanisms for models with and without TaSc (No-TaSc). \underline{Underlined} and {\bf bold} values indicate comparable and better predictive performance by using TaSc respectively. Standard deviations do not exceed 0.01}
\label{tab:data_performances}
\end{table}

\footnotetext{Lower predictive performance is observed with BERT in MIMIC, as BERT accepts a maximum of 512 word pieces as input. See Appendix \ref{sec:appendix_model_hyperparameters}.}

\subsection{Decision Flip: Most Informative Token}
\label{sec:single_attention_weight}

Tables \ref{tab:decision_single}, \ref{tab:decision_single_all_encoders} and \ref{tab:decision_single_all_dataset} present the mean average percentage of decision flips (higher is better) across attention mechanisms, encoders and datasets by removing the most informative token for TaSc variants and No-TaSc for all attention-based importance metrics (see Section~\ref{sec:interpretability}).

\begin{table}[!ht]
\setlength\tabcolsep{1.2pt}
\def\arraystretch{1}
\small
\centering
\begin{tabular}{c | l|| c |r|r|r  }
 & \textbf{Att.}& \textbf{No-TaSc} &\textbf{Lin-TaSc} & \textbf{Feat-TaSc} & \textbf{Conv-TaSc} \\ \hline \hline

\multirow{2}{*}{$\alpha$} 
& \textbf{Tanh} &     \textbf{8.4} &   7.3 (0.9) &   6.5 (0.8) &   5.4 (0.6) \\
& \textbf{Dot} &     \textbf{5.4} &   4.3 (0.8) &   4.8 (0.9) &   4.5 (0.8) \\ \hline
\multirow{2}{*}{$\nabla\alpha$} 
& \textbf{Tanh} &     8.2 &  10.2 (1.2) &  \textbf{11.2} (1.4) &  10.4 (1.3) \\
& \textbf{Dot} &     6.9 &  10.9 (1.6) &   \textbf{12.2} (1.8) &  11.1 (1.6) \\ \hline
\multirow{2}{*}{$\alpha\nabla \alpha$} 
& \textbf{Tanh}  &    11.7 &  \textbf{\underline{14.0}} (1.2) &  13.5 (1.1) &  12.2 (1.0) \\
& \textbf{Dot} &     8.2 &  11.8 (1.4) &  \textbf{\underline{12.6}} (1.5) &  11.3 (1.4) 

\end{tabular}
\caption{Mean average \textit{percentage of decision flips} across attention mechanisms occurred by removing the most informative token, using the three TaSc variants and No-TaSc (higher is better). \textbf{Bold} and \underline{underlined} values denote best performing method row-wise and overall (for each attention mechanism). Relative improvement over No-TaSc in parenthesis ($>$1 TaSc is better than No-TaSc).}
\label{tab:decision_single}
\end{table}


\begin{table}[!ht]
\setlength\tabcolsep{2pt}
\def\arraystretch{1}
\small
\centering
\begin{tabular}{c | l||c|c|c |c }
 & \textbf{Enc()}& \textbf{No-TaSc} &\textbf{Lin-TaSc} & \textbf{Feat-TaSc} & \textbf{Conv-TaSc}  \\ \hline \hline

\multirow{5}{*}{$\alpha$} 
&  \textbf{BERT} &  4.8 &   6.2 (1.3) &   \textbf{6.6} (1.4) &   3.7 (0.8) \\
&  \textbf{LSTM} &  \textbf{6.2 }&   4.8 (0.8) &   5.1 (0.8) &   4.8 (0.8) \\
&   \textbf{GRU} &  \textbf{6.2} &   5.7 (0.9) &   5.5 (0.9) &   5.4 (0.9) \\
&   \textbf{MLP} &  \textbf{8.0 }&   6.0 (0.8) &   5.2 (0.7) &   5.6 (0.7) \\
&   \textbf{CNN} &  \textbf{9.3} &   6.2 (0.7) &   5.7 (0.6) &   5.4 (0.6) \\\hline

\multirow{5}{*}{$\nabla \alpha$} 
&  \textbf{BERT} &  5.2 &   6.4 (1.2) &   \textbf{7.4} (1.4) &   3.6 (0.7) \\
&  \textbf{LSTM} &  6.5 &  10.9 (1.7) &  \textbf{12.5} (1.9) &  12.0 (1.9) \\
&   \textbf{GRU} &  6.3 &  11.3 (1.8) &  \textbf{12.4} (2.0) &  11.8 (1.9) \\
&   \textbf{MLP} &  9.8 &  12.0 (1.2) &  \textbf{13.1 }(1.3) &  13.0 (1.3) \\
&   \textbf{CNN} & 10.1 &  12.0 (1.2) &  \textbf{13.2} (1.3) &  13.4 (1.3) \\ \hline

\multirow{5}{*}{$\alpha \nabla \alpha$} 
&  \textbf{BERT} &  5.7 &   8.0 (1.4) &   \textbf{\underline{9.3}} (1.6) &   4.0 (0.7) \\
&  \textbf{LSTM} &  8.3 &  12.8 (1.5) &  \textbf{\underline{13.6}} (1.6) &  13.1 (1.6) \\
&   \textbf{GRU} &  8.3 &  \textbf{\underline{14.2}} (1.7) &  13.9 (1.7) &  13.4 (1.6) \\
&   \textbf{MLP} & 13.7 &  \textbf{\underline{14.6}} (1.1) &  13.9 (1.0) &  13.8 (1.0) \\
&   \textbf{CNN} & 13.9 &  \textbf{\underline{14.7}} (1.1) &  14.5 (1.0) &  14.6 (1.0) 
\end{tabular}
\caption{Mean average \textit{percentage of decision flips} occurred by removing the most informative token, using the three TaSc variants and No-TaSc across encoders (higher is better). }
\label{tab:decision_single_all_encoders}
\end{table}

In Table~\ref{tab:decision_single}, we observe that TaSc variants are effective in identifying the single most important token, outperforming No-TaSc in 12 out of 18 cases across attention-based importance metrics. This suggests that the attention mechanisms benefit from the non-contextualised information encapsulated in TaSc when allocating importance to the input tokens. Models using Tanh without TaSc appear to produce on average a higher percentage of decision flips compared to those using the Dot mechanism. Using either of the TaSc variants improves both mechanisms, with Dot mechanism benefiting the most, making it comparable to Tanh. For example, Dot moves from 8.2\% with No-TaSc to 11.8\% with Lin-TaSc, which is closer to 14.0\% achieved by Lin-TaSc with Tanh (for $\alpha \nabla \alpha$).

\begin{table}[!t]
\setlength\tabcolsep{2pt}
\def\arraystretch{0.9}
\small
\centering
\begin{tabular}{c | l||c|c|c |c  }
 & \textbf{Dataset}& \textbf{No-TaSc} &\textbf{Lin-TaSc} & \textbf{Feat-TaSc} & \textbf{Conv-TaSc}  \\ \hline \hline

\multirow{5}{*}{$\alpha$} 
&    \textbf{SST} & \textbf{16.7} &  12.5 (0.7) &  11.2 (0.7) &  11.5 (0.7) \\
&    \textbf{ADR} &  4.5 &   \textbf{5.8} (1.3) &   5.4 (1.2) &   3.9 (0.9) \\
&   \textbf{IMDB} &  \textbf{6.6} &   5.0 (0.8) &   5.2 (0.8) &   3.9 (0.6) \\
&     \textbf{AG} &  3.5 &   3.1 (0.9) &   \textbf{3.8} (1.1) &   2.7 (0.8) \\
&  \textbf{MIMIC} &  \textbf{3.0} &   2.5 (0.8) &   2.5 (0.8) &   2.5 (0.8) \\ \hline

\multirow{5}{*}{$\nabla \alpha$} 
&    \textbf{SST} & 18.8 &  25.8 (1.4) &  \textbf{27.5} (1.5) &  25.4 (1.3) \\
&    \textbf{ADR} &  5.2 &   9.4 (1.8) &  \textbf{10.6} (2.0) &   8.1 (1.5) \\
&   \textbf{IMDB} &  7.1 &   7.6 (1.1) &   9.2 (1.3) &   \textbf{9.3} (1.3) \\
&     \textbf{AG} &  4.3 &   5.1 (1.2) &   6.4 (1.5) &   6.4 (1.5) \\
&  \textbf{MIMIC} &  2.5 &   \textbf{4.8} (1.9) &   \textbf{4.8} (1.9) &   \textbf{4.8} (1.9) \\ \hline

\multirow{5}{*}{$\alpha \nabla \alpha$} 
&    \textbf{SST} & 24.1 &  29.4 (1.2) &  \textbf{\underline{29.5}} (1.2) &  27.2 (1.1) \\
&    \textbf{ADR} &  6.0 &  10.3 (1.7) &  \textbf{\underline{11.1}} (1.8) &   8.2 (1.4) \\
&   \textbf{IMDB} & 10.3 &  \textbf{\underline{12.0}} (1.2) &  11.0 (1.1) &  10.8 (1.0) \\
&     \textbf{AG} &  5.2 &   6.4 (1.2) &   \textbf{\underline{8.0}} (1.5) &   6.9 (1.3) \\
&  \textbf{MIMIC} &  4.3 &   \textbf{\underline{6.2}} (1.5) &   5.7 (1.3) &   5.8 (1.3) \\

\end{tabular}
\caption{Mean average \textit{percentage of decision flips} occurred by removing the most informative token, using the three TaSc variants and No-TaSc across datasets (higher is better).} 
\label{tab:decision_single_all_dataset}
\end{table}

Table~\ref{tab:decision_single_all_encoders} presents a comparison across encoders. TaSc variants achieve improved performance over No-TaSc in 30 out 45 cases. All TaSc variants yield comparable results with the exception of Conv-TaSc with BERT. Results further suggest that non-recurrent encoders (MLP, CNN) without TaSc outperform recurrent encoders (LSTM, GRU) and BERT which has the poorest performance. We hypothesise that this is due to the attention module becoming more important without feature contextualisation which is similar to findings of \citet{serrano2019attention} and \citet{wiegreffe2019attention}. However, we observe that using any of the TaSc variants across encoders results into improvements with LSTM and GRU becoming comparable to MLP and CNN. For example, BERT without TaSc improves from 5.7\% to 8.0\% (relative improvement 1.4x) and 9.3\% (relative improvement 1.6x) using Lin-TaSc and Feat-TaSc respectively (for $\alpha \nabla \alpha$).

In Table~\ref{tab:decision_single_all_dataset}, we see that TaSc variants outperform No-TaSc in 33 out of 45 cases across datasets. This highlights the robustness of TaSc as improvements are irrespective of the dataset. In general, Lin-TaSc and Feat-TaSc perform equally well, however Lin-TaSc has the smaller number of parameters amongst the three variants. Similar to the findings of \citet{serrano2019attention} best results overall, irrespective of the use of TaSc, are obtained using $\alpha \nabla \alpha$ to rank importance.

\subsection{Decision Flip: Fraction of Tokens}

Providing one token (i.e., the most informative) as an explanation is not always a realistic approach to assessing faithfulness. In our second experiment, we test TaSc by measuring the fraction of important tokens required to be removed to cause a decision flip (change model's prediction). Tables \ref{decision_set_all}, \ref{tab:decision_set_all_encoders} and \ref{tab:decision_set_all_datasets} show the mean average fraction of tokens required to be removed to cause a decision flip (lower is better) across attention mechanisms, encoders and datasets for all importance metrics.

\begin{table}[!ht]
\setlength\tabcolsep{1.2pt}
\def\arraystretch{1}
\small
\centering
\begin{tabular}{c | l|| c | c|c|c  }
 & \textbf{Att.}& \textbf{No-TaSc} & \textbf{Lin-TaSc} & \textbf{Feat-TaSc} & \textbf{Conv-TaSc} \\ \hline \hline

\multirow{2}{*}{$\alpha$} 
& \textbf{Tanh} & .44 &  \textbf{.39} (0.9) &  .42 (0.9) &  .43 (1.0) \\ 
& \textbf{Dot} & .60 &   \textbf{.52} (0.9) &  .53 (0.9) &  .56 (0.9) \\ \hline

\multirow{2}{*}{$\nabla \alpha$} 
& \textbf{Tanh} & .36 &  .21 (0.6) &  \textbf{.19} (0.5) &  .26 (0.7) \\ 
& \textbf{Dot} & .42 &  \textbf{.22} (0.5) &   \textbf{.22} (0.5) &  .26 (0.6) \\ \hline

\multirow{2}{*}{$\alpha \nabla \alpha$} 
& \textbf{Tanh} & .32 &  \textbf{\underline{.17}} (0.5) &  .18 (0.5) &  .24 (0.7) \\ 
& \textbf{Dot} &  .41 &  \textbf{\underline{.21}} (0.5) &   \textbf{\underline{.21}} (0.5) &  .26 (0.6) \\ 

\end{tabular}
\caption{Mean average \textit{fraction of informative tokens} required to cause a decision flip across attention mechanisms, using the three TaSc variants and No-TaSc (lower is better). \textbf{Bold} and \underline{underlined} values denote best performing method row-wise and overall (for each attention mechanism). Relative improvement over No-TaSc in parenthesis ($<$1 TaSc is better than No-TaSc).}
\label{decision_set_all}
\end{table}

\begin{table}[!ht]
\setlength\tabcolsep{2pt}
\def\arraystretch{1}
\small
\centering
\begin{tabular}{c | l|| c |c|c|c }
& \textbf{Enc()}&\textbf{No-TaSc}  & \textbf{Lin-TaSc} & \textbf{Feat-TaSc} & \textbf{Conv-TaSc}   \\ \hline \hline

\multirow{5}{*}{$\alpha$} 
&  \textbf{BERT} &  .59 &  .46 (0.8) &  \textbf{.44} (0.7) &  .56 (0.9) \\
&  \textbf{LSTM} &  .56 &  \textbf{.48} (0.9) &  .51 (0.9) &  .52 (0.9) \\
&   \textbf{GRU} &  .57 &  \textbf{.45}  (0.8) &  .49 (0.9) &  .50 (0.9) \\
&   \textbf{MLP} &  \textbf{.41} &  .43 (1.0) &  .44 (1.1) &  .46 (1.1) \\
&   \textbf{CNN} &  .45 &  .47 (1.0) &  .47 (1.0) &  \textbf{.44} (1.0)  \\ \hline

\multirow{5}{*}{$\nabla \alpha$} 
&  \textbf{BERT} &  .52 &  .34 (0.6) &  \textbf{.31} (0.6) &  .58 (1.1) \\
&  \textbf{LSTM} &  .44 &  .21 (0.5) &  \textbf{.17} (0.4) &  .19 (0.4) \\
&   \textbf{GRU} &  .46 &  \textbf{.17} (0.4) &  .18 (0.4) &  .19 (0.4) \\
&   \textbf{MLP} &  .21 &  .18 (0.8) &  \textbf{.17} (0.8) &  \textbf{.17} (0.8) \\
&   \textbf{CNN} &  .30 &  \textbf{.17} (0.6) &  \textbf{.17} (0.6) &  \textbf{.17} (0.6) \\\hline

\multirow{5}{*}{$\alpha \nabla \alpha$} 
&  \textbf{BERT} &  .52 &   \textbf{\underline{.29}} (0.6) &  \textbf{\underline{.29}} (0.6) &  .57 (1.1) \\
&  \textbf{LSTM} &  .44 &   .20 (0.5) &  \textbf{\underline{.16}} (0.4) &  .18 (0.4) \\
&   \textbf{GRU} &  .43 &  \textbf{\underline{.16}} (0.4) &  .17 (0.4) &  .18 (0.4) \\
&   \textbf{MLP} &  .17 &  \textbf{\underline{.15}} (0.9) &  .16 (0.9) &  .16 (0.9) \\
&   \textbf{CNN} &  .27 &  \textbf{\underline{.16}} (0.6) &  \textbf{\underline{.16}}(0.6) &  \textbf{\underline{.16}} (0.6) \\

\end{tabular}
\caption{Mean average \textit{fraction of tokens} required to cause a decision flip, using the three TaSc variants and No-TaSc across encoders (lower is better). }

\label{tab:decision_set_all_encoders}

\end{table}



\begin{table}[!t]
\setlength\tabcolsep{2pt}
\def\arraystretch{1}
\small
\centering
\begin{tabular}{c | l|| c |c|c|c }
&   \textbf{Dataset}& \textbf{No-TaSc} &\textbf{Lin-TaSc} & \textbf{Feat-TaSc} & \textbf{Conv-TaSc} \\ \hline \hline

\multirow{5}{*}{$\alpha$} 
&    \textbf{SST} &  \textbf{.45} &  .49 (1.1) &  .48 (1.1) &   .50 (1.1) \\
&    \textbf{ADR} &  .88 &  \textbf{.72} (0.8) &  .76 (0.9) &  .77 (0.9) \\
&   \textbf{IMDB} &  .38 &  \textbf{.31} (0.8) &  .36 (1.0) &  .41 (1.1) \\
&     \textbf{AG} &  .59 &  .52 (0.9) &  \textbf{.49} (0.8) &  .54 (0.9) \\
&  \textbf{MIMIC} &  .28 &  \textbf{.24} (0.9) &  .26 (0.9) &  .26 (0.9) \\ \hline

\multirow{5}{*}{$\nabla \alpha$} 
&    \textbf{SST} &  .35 &  .24 (0.7) &  \textbf{.20} (0.6) &  .27 (0.8) \\
&    \textbf{ADR} &  .78 &  .38 (0.5) &  \textbf{.36} (0.5) &  .47 (0.6) \\
&   \textbf{IMDB} &  .19 &  \textbf{.10} (0.5) &  .12 (0.6) &  .14 (0.7) \\
&     \textbf{AG} &  .49 &  .35 (0.7) &  \textbf{.28} (0.6) &  .35 (0.7) \\
&  \textbf{MIMIC} &  .13 &  \textbf{\underline{.02}} (0.2) &  .03 (0.3) &  .07 (0.5) \\ \hline

\multirow{5}{*}{$\alpha \nabla \alpha$} 
&    \textbf{SST} &  .33 &  .23 (0.7) &  \textbf{\underline{.19}} (0.6) &  .26 (0.8) \\
&    \textbf{ADR} &  .77 &  \textbf{\underline{.34}} (0.5) &  .35 (0.5) &  .47 (0.6) \\
&   \textbf{IMDB} &  .17 &  \textbf{\underline{.07}} (0.4) &  .10 (0.6) &  .13 (0.7) \\
&     \textbf{AG} &  .46 &  .31 (0.7) &  \textbf{\underline{.25}} (0.5) &  .34 (0.7) \\
&  \textbf{MIMIC} &  .10 &  \textbf{\underline{.02}} (0.2) &  .03 (0.3) &  .06 (0.7) \\

\end{tabular}
\caption{Mean average \textit{fraction of tokens} required to cause a decision flip, using the three TaSc variants and No-TaSc across datasets (lower is better).}
\label{tab:decision_set_all_datasets}

\end{table}

In Table \ref{decision_set_all}, we see that attention-based explanations from models trained with any of the TaSc mechanisms require on average a lower fraction of tokens to cause a decision flip compared to No-TaSc (in 17 out of 18 cases). Overall Lin-TaSc achieves higher or comparable relative improvements over Conv-TaSc and Feat-TaSc in 5 out of 6 times.

Table~\ref{tab:decision_set_all_encoders} shows results across encoders. All three TaSc variants obtain comparable performance with the exception of Conv-TaSc with BERT. We hypothesise that with BERT, Conv-TaSc fails to capture interactions between embedding dimensions due to perhaps higher contextualisation of BERT embeddings (i.e. contain more duplicate information). Similarly to the previous experiment results suggest that non-recurrent encoders (MLP and CNN) without TaSc outperform the remainder of encoders, with BERT having the worst performance. This strengthens our hypothesis that attention becomes more important to a model with reduced contextualisation. When using TaSc, performance across all encoders becomes comparable with the exception of BERT. For example, GRU improves from .43 with No-TaSc to .16 with Lin-TaSc, .17 with Feat-TaSc and .18 with Conv-TaSc (for $\alpha \nabla \alpha$).

Table~\ref{tab:decision_set_all_datasets} presents results across datasets. All three TaSc mechanims manage to outperform vanilla attention. Lin-TaSc and Feat-TaSc perform comparably, with the first having a slight edge obtaining highest relative improvements in 3 out of 5 datasets with $\alpha \nabla \alpha$. For example in ADR, No-TaSc requires on average .77 of all tokens to be removed for a decision flip to occur compared to .34 obtained by Lin-TaSc (for $\alpha \nabla \alpha$). The benefits of TaSc become evident when considering longer sequences. For example in MIMIC, Lin-TaSc requires on average 44 tokens to cause a decision flip compared to 220 for No-TaSc.

\subsection{Robustness Analysis}

\begin{figure}[!ht]
  \centering
  \includegraphics[height=75mm,width=\linewidth]{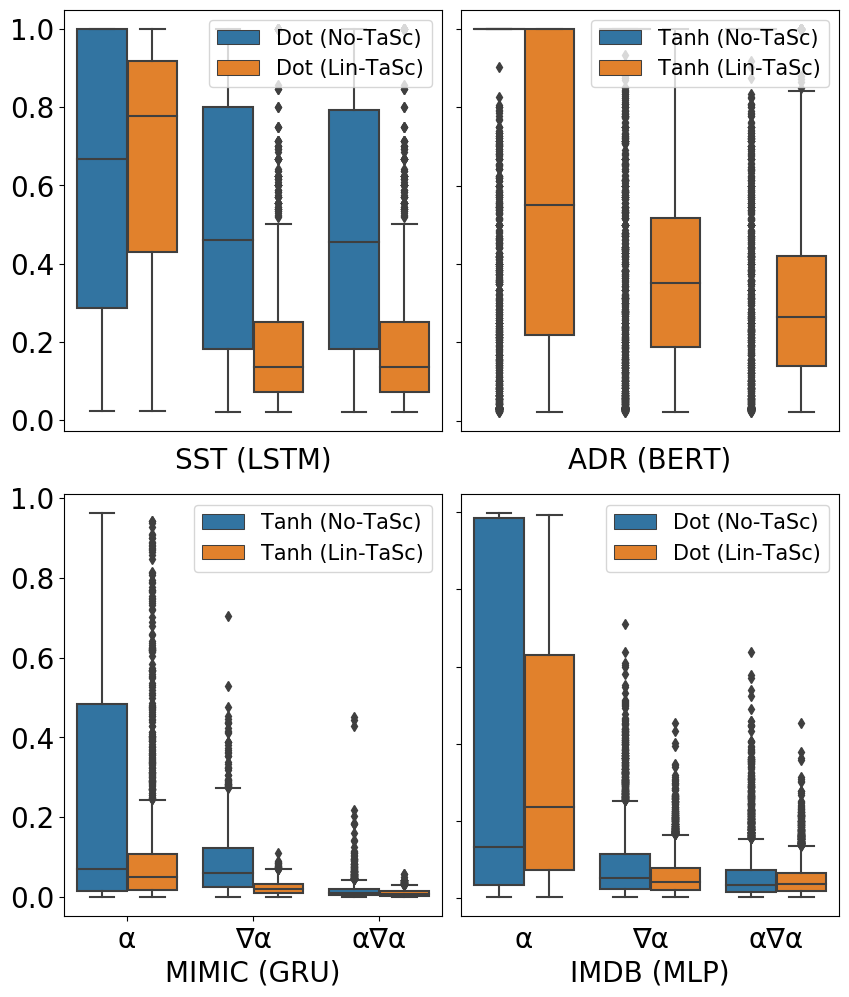}
  \caption{Box-plots of \textit{fractions of tokens} removed across all test instances and importance metrics. \;  \put(1,3){\circle{3}}\put(1,3){\color{blue}\circle*{7}} \; denotes attention without TaSc; \;\; \put(1,3){\circle{3}}\put(1,3){\color{orange}\circle*{7}} \;\; denotes attention with Lin-TaSc (lower and narrower is better).}
  \label{fig:decision_flip_set}
\end{figure}

We also perform a detailed comparison between the best performing TaSc variant (Lin-TaSc) and vanilla attention (No-TaSc) across all test instances. Figure \ref{fig:decision_flip_set} shows box-plots with the median fraction of tokens required to be removed for causing a decision flip when ranking tokens by all three importance metrics. For brevity we present results for four cases.

We notice that the median fraction of tokens required to cause a decision flip for Lin-TaSc using $\alpha$ is higher compared to No-TaSc in certain cases. However, Lin-TaSc results in consistently lower medians (with substantially reduced variances) compared to No-TaSc using $\nabla \alpha$ and $\alpha \nabla \alpha$ which are more effective importance metrics. This is particularly visible in the ADR dataset with BERT, where the 25\% and 75\% percentiles are much closer to the median values, compared to No-TaSc. Reduced variances suggest that the explanation faithfulness across instances remains consistent.

\section{Comparing TaSc with Non-attention Input Importance Metrics}
\label{sec:results_tasc_vs_interpretability}



We finally compare explanations provided by using Lin-TaSc and $\alpha \nabla \alpha$ to three standard non-attention input importance metrics without TaSc which are strong baselines for explainability~\citep{nguyen2018comparing,atanasova2020diagnostic}.


\paragraph{\textbf{Word Omission (\textbf{WO})} \citep{robnik2008explaining,nguyen2018comparing}:} Ranking input words by computing the difference between the probabilities of the predicted class when including a word $i$ and omitting it: $\text{WO}_i = p(\hat{y}|\mathbf{x}) - p(\hat{y}|\mathbf{x}_{\backslash x_i}) $

\paragraph{\textbf{InputXGrad ($\mathbf{x} \nabla \mathbf{x}$)} \citep{kindermans2016investigating,atanasova2020diagnostic}:} Ranking words by multiplying the gradient of the input by the input with respect to the predicted class: $\nabla x_i = \frac{\partial \hat{y}}{\partial x_i}$

\paragraph{\textbf{Integrated Gradients ($\mathbf{IG}$)} \citep{integrated_gradients}:} Ranking words by computing the integral of the gradients taken along a straight path from a baseline input to the original input, where the baseline is the zero embedding vector.

\paragraph{Comparison Results}
Table \ref{tab:tasc_gradient_comparison} shows the results on decision flip (fraction of tokens removed) comparing the best performing attention-based importance metric ($\alpha \nabla \alpha$) with Lin-TaSc to Non-TaSc models with WO, $\mathbf{x} \nabla \mathbf{x}$ and IG importance metrics across all encoders and datasets.\footnote{We do not compare with LIME \citep{ribeiro2016model} because WO and the gradient-based approaches outperform it \citep{nguyen2018comparing,atanasova2020diagnostic}.} We observe that using $\alpha \nabla \alpha$ with TaSc to rank word importance requires a lower fraction of tokens to cause a decision flip on average compared to WO, $\mathbf{x} \nabla \mathbf{x}$ and IG without TaSc. We outperform the other explanation approaches in 40 out of 50 cases, whilst obtaining comparable performance in other 5 cases. This demonstrates the efficacy of TaSc in providing more faithful attention-based explanations than strong baselines without TaSc~\citep{nguyen2018comparing,atanasova2020diagnostic}.
The improvements are particularly evident using BERT as an encoder. In IMDB, WO with Tanh requires on average .23 of the tokens to be removed for a decision flip compared to just .07 for $\alpha \nabla \alpha$ with TaSc. 


We also observe that the attention-based importance metric ($\alpha \nabla \alpha$) with TaSc is a more robust explanation technique than non-attention based ones, obtaining lower variance in the  fraction of tokens required to cause a decision flip across encoders. For example $\alpha \nabla \alpha$ with TaSc and Tanh requires a fraction of tokens in the range of .01-.05 compared to IG which requires .02-.43 in MIMIC, showing the consistency of our proposed approach.



Finally we observe that TaSc consistently improves non-attention based explanation approaches (WO, $\mathbf{x} \nabla \mathbf{x}$ and IG) requiring a lower fraction of tokens to be removed compared to Non-TaSc across encoders, datasets and attention mechanisms in the majority of cases (see full results in Appendix \ref{appendix:comparing_tasc}).

\begin{table}[!t]
    \small
    \centering
    \setlength\tabcolsep{1.45pt}
\begin{tabular}{cc||cccc||cccc}
                      &         & \multicolumn{4}{c||}{\textbf{Tanh}}                           & \multicolumn{4}{c}{\textbf{Dot}}                                     \\
                      &         & \multicolumn{3}{c}{\textbf{Non-TaSc}} & \textbf{TaSc}              & \multicolumn{3}{c}{\textbf{Non-TaSc}}            & \textbf{TaSc}              \\
\textbf{Data }               & \textbf{Enc()} & \textbf{WO}   & $\mathbf{x} \nabla \mathbf{x} $   & \textbf{IG}     & $\alpha \nabla \alpha$ & $\textbf{WO} $   & $\mathbf{x} \nabla \mathbf{x} $   & \textbf{IG}     & $\alpha \nabla \alpha$ \\ \hline
\multirow{5}{*}{SST}   & BERT &            .29 &        .64 &      .51 &                    \textbf{.22} &           \textbf{.32} &       .62 &     .49 &                   .55 \\
          &    LSTM &            .25 &        .24 &      .20 &                    \textbf{.19} &           .21 &       .23 &     \textbf{.19 }&                   \textbf{.19 }\\
          &     GRU &            .24 &        .22 &      .19 &                    \textbf{.18} &           .24 &       .25 &     .23 &                   \textbf{.19} \\
          &     MLP &            .36 &        .26 &      .24 &                    \textbf{.18} &           .22 &       .19 &     \textbf{.18 }&                   \textbf{.18} \\
          &     CNN &            .30 &        .25 &      .20 &                    \textbf{.19} &           .22 &       .20 &     \textbf{.18} &                   .19 \\ \hline
\multirow{5}{*}{ADR}   & BERT &            .83 &        .91 &      .89 &                    \textbf{.31} &           .81 &       .90 &     .87 &                   \textbf{.50} \\
      &    LSTM &            .82 &        .81 &      .80 &                    \textbf{.32} &           .87 &       .88 &     .87 &                   \textbf{.34} \\
      &     GRU &            .84 &        .84 &      .84 &                    \textbf{.35} &           .79 &       .80 &     .80 &                   \textbf{.38} \\
      &     MLP &            .71 &        .63 &      .57 &                    \textbf{.31} &           .49 &       .43 &     \textbf{.39} &                   .40 \\
      &     CNN &            .80 &        .78 &      .78 &                    \textbf{.37} &           .77 &       .74 &     .74 &                   \textbf{.36} \\ \hline
\multirow{5}{*}{IMDB}  & BERT &            .23 &        .69 &      .43 &                    \textbf{.07} &           .24 &       .72 &     .49 &                   \textbf{.20} \\
         &    LSTM &            .18 &        .12 &      .07 &                    \textbf{.04} &           .26 &       .09 &     .07 &                   \textbf{.05} \\
         &     GRU &            .18 &        .12 &      .07 &                    \textbf{.04} &           .27 &       .15 &     .08 &                   \textbf{.05} \\
         &     MLP &            .16 &        \textbf{.05} &      \textbf{.05} &                    \textbf{.05} &           .18 &       .07 &     .06 &                   \textbf{.05} \\
         &     CNN &            .21 &        .09 &      .07 &                    \textbf{.05 }&           .27 &       .07 &     .06 &                   \textbf{.05} \\ \hline
\multirow{5}{*}{AG}    & BERT &            .62 &        .78 &      .56 &                    \textbf{.50} &           .56 &       .76 &     \textbf{.60} &                   \textbf{.60} \\
      &    LSTM &            .53 &        .51 &      \textbf{.30} &                    .38 &           .47 &       .52 &     \textbf{.35} &                   .46 \\
      &     GRU &            .45 &        .36 &      .31 &                    \textbf{.20} &           .54 &       .40 &     .30 &                   \textbf{.22} \\
      &     MLP &            .53 &        .24 &      .25 &                    \textbf{.19} &           .44 &       .25 &     .23 &                   \textbf{.19} \\
      &     CNN &            .55 &        .38 &      .28 &                    \textbf{.20} &           .53 &       .35 &     .25 &                   \textbf{.21 }\\ \hline
\multirow{5}{*}{MIMIC} & BERT &            .24 &        .67 &      .43 &                    \textbf{.03} &           .21 &       .57 &     .26 &                   \textbf{.05} \\
  &    LSTM &            .35 &        .32 &      .12 &                    \textbf{.01} &           .28 &       .40 &     .30 &                   \textbf{.01} \\
  &     GRU &            .20 &        .24 &      .23 &                    \textbf{.01} &           .36 &       .18 &     .08 &                   \textbf{.01} \\
  &     MLP &            .40 &        .03 &      .22 &                    \textbf{.01} &           .13 &       .04 &     .03 &                   \textbf{.02} \\
  &     CNN &            .26 &        .15 &      .02 &                    \textbf{.01} &           .43 &       .09 &     \textbf{.02} &                   \textbf{.02} \\
\end{tabular}
\caption{Average \textit{fraction of tokens} required to cause a decision flip using the best performing attention-based ranking ($\alpha \nabla \alpha$) with TaSc, \textit{Word omission} without TaSc (\textbf{WO}), \textit{InputXGrad} without TaSc ($\nabla \mathbf{x}$) and \textit{Integrated Gradients} without TaSc ($\mathbf{IG}$).}
\label{tab:tasc_gradient_comparison}
\end{table}

 \section{Conclusion}

We introduced TaSc, a family of three encoder-independent mechanisms that induce context-independent task-specific information to attention. We conducted an extensive series of experiments showing the superiority of TaSc over vanilla attention on improving faithfulness of attention-based interpretability without sacrificing predictive performance. Finally, we show that attention-based explanations with TaSc outperform other interpretability techniques. For future work, we will explore the effectiveness of TaSc in sequence-to-sequence tasks similar to \citet{Vashishth2019} and explore the inner attention mechanisms of BERT.

\section*{Acknowledgments}
We would like to thank the anonymous reviewers for their constructive and detailed comments that helped to improve the paper through revise and resubmit. Nikolaos Aletras is supported by EPSRC grant EP/V055712/1, part of the European Commission CHIST-ERA programme, call 2019 XAI: Explainable Machine Learning-based Artificial Intelligence.

\bibliographystyle{acl_natbib}
\bibliography{acl2021}


\clearpage
\newpage
\begin{appendices}

\section{Model Hyperparameters \label{sec:appendix_model_hyperparameters}}

Similar to \citet{jain2019attention} we use FastText pretrained embeddings \citep{fastext} for the SST and ADR 
datasets, Glove pretrained embeddings \citep{pennington-etal-2014-glove} for the IMDB and AG News datasets, while we use Word2Vec \citep{mikolov} from Gensim \citep{rehurek_lrec} to train embeddings for MIMIC.
All embeddings are of size $d$ = 300. We also replace all numbers in text with a special symbol $q$ and initialise the embeddings of unknown words randomly from a normal distribution, $\mathcal{N}(0,1)$. The embeddings are not trained alongside the rest of the model.

We train the models using default Adam learning rate (1e-3) with 1e-4 weight decay, which adds an $l_2$ regulariser across all parameters. We use 64 dimensional hidden representations for one-layered bi-LSTM and bi-GRU encoders and 128 dimensional hidden representation for the MLP encoder following \citet{jain2019attention}. For the CNN we use 4 kernels of sizes [1, 3, 5, 7], each with 32 filters, giving a final contextual representation $\mathbf{h}_i$ of size $N = 128$, with ReLU activation function on the output of the filters, as per \citet{jain2019attention}. 

For BERT we use the pre-trained version from \citet{Wolf2019HuggingFacesTS} and fine-tune with a learning rate of $1e-5$ all BERT parameters except from the word embeddings, to simulate the scenario with the rest of the encoders, and $1e-4$ for the remainder of the parameters. We train our models three times using different random seeds and a batch size of 8 for BERT and 32 for the rest of the models.

For Conv-TaSc we apply a CNN with 15 channels over the scaled embedding $\mathbf{e}_i$ from Lin-TaSc, keeping a single stride and a 1-dimensional kernel. This way, we ensure that input words remain context-independent. We then sum over the filtered scaled embedding $\mathbf{e}^f_i$, to obtain the scores $s_{x_i}$. We have also experimented with filter sizes of [2, 10, 20 , 30, 50] individually and simultaneously. 

For the MIMIC dataset we also attempted to use LongFormer \citep{beltagy2020longformer}, which is a BERT version that has the ability to accept and deal with longer sequences. However due to the increasing time to train and evaluate the model, this BERT variant was abandoned. Additionally we attempted to use Hierarchical BERT to deal with the longer sequences, however increases where not substantial and run times where similarly increased. Finally, contrary to the remainder of the datasets to deal with the long sequences of MIMIC we truncated the 256 first tokens and 256 last tokens, following the suggestions of \citet{sun2019fine}. We experimented with using the first and the last 512 tokens, but the head and tails truncation approach yielded the best performances.

\section{Additional parameters with TaSc variants \label{sec:additional_parameters}}

In Table \ref{tab:additional_params} we present the additional parameters introduced by each variant, with Lin-TaSc requiring the lowest number of parameters and Feat-TaSc the most.

\begin{table}[!ht]
    \centering
    \begin{tabular}{c|c}
        \textbf{TaSc Mechanism} &  \textbf{Additional Parameters}  \\ \hline
        Lin-TaSc & $\mathbf{|V|}$  \\
        Feat-TaSc & $\mathbf{|V|} \times d$\\
        Conv-TaSc & $\mathbf{|V|} + d \times n + n$
    \end{tabular}
    \caption{Additional parameters resulting from the proposed TaSc mechanisms where $|V|$ is the vocabulary size, $d$ the embedding dimension and $n$ the number of channels in a CNN. }
    \label{tab:additional_params}
\end{table}

\section{Reproducibility Results}

\paragraph{\textbf{Computational infrastructure used:}} For the experiments above we used NVIDIA's TESLA V100 GPU.

\paragraph{\textbf{Dataset description:}} We consider the following datasets for text classification following~\citet{wiegreffe2019attention} and \citet{jain2019attention}:
\paragraph{SST:} \textit{Stanford Sentiment Treebank}  consists of sentences tagged with sentiment on a 5-point-scale from negative to postive \citep{socher2013sst}. \citet{jain2019attention} removed sentences with neutral sentiment and labelled the remaining sentences to negative and positive if they have a score lower or higher than 3 respectively.

\paragraph{IMDB:} The \textit{Large Movie Reviews Corpus} consists of 50,000 movie reviews labelled either as positive or negative \citep{maas2011learning}. We filter the dataset as per \citet{jain2019attention} to include movie reviews with sequence length less than 400 words.

\paragraph{ADR:} A dataset of $\sim$20,000 tweets with labels indicating whether a Twitter post contains an adverse drug reaction or not \cite{sarker2015utilizing}. 

\paragraph{AG:} A subset of the original news articles\footnote{\url{https://di.unipi.it/~gulli/AG_corpus_of_news_articles.html}. Accessed on Sep 2019} dataset compiled by \citet{jain2019attention} for topic categorisation (\textit{Business} and \textit{World} news). 

\paragraph{MIMIC:} A sample of discharge summaries from the MIMIC III dataset of health records \citep{johnson2016mimic}. The task is to recognise if a given summary has been labelled as relevant to acute or chronic anemia \citep{jain2019attention}. 

\paragraph{\textbf{Validation set predictive performances: }} In Table \ref{tab:data_performances_validation} we present predictive performances on the validation checks for reproducibility on models with TaSc and models without (No-TaSc).

\begin{table}[!t]
\setlength\tabcolsep{1pt}
\small
\centering
\begin{tabular}{cc|cc|cc|cc|cc}
\multicolumn{1}{c}{\textbf{Data-}}              & \multicolumn{1}{c|}{\textbf{Enc()}}             & \multicolumn{2}{c|}{\textbf{No-TaSc}}                         & \multicolumn{2}{c|}{\textbf{Lin-TaSc}}                                       & \multicolumn{2}{c|}{\textbf{Feat-TaSc}}                                                                          & \multicolumn{2}{c}{\textbf{Conv-TaSc}}                                                                         \\
\textbf{set}& \textbf{} & \textbf{Dot} & \textbf{Tanh} & \textbf{Dot} & \textbf{Tanh}   &  \textbf{Dot} & \textbf{Tanh} &  \textbf{Dot} & \textbf{Tanh} \\ \hline
\multirow{5}{*}{SST}   

&     BERT &  .89 &   .90 &  .90 &  .87 &  .87 &  .87 &  .90 &  .90 \\
&     LSTM &  .77 &  .78 &  .77 &  .78 &  .77 &   .80 &  .79 &   .80 \\
&      GRU &  .78 &  .78 &  .78 &  .79 &  .78 &  .79 &  .78 &  .79 \\
&   MLP    &  .75 &  .77 &  .78 &  .78 &   .80 &   .80 &  .79 &  .81 \\
&      CNN &  .77 &  .77 &  .79 &   .80 &   .80 &  .79 &  .79 &  .78 \\ \hline

\multirow{5}{*}{ADR}             
&     BERT &  .81 &  .81 &  .81 &  .79 &   .80 &   .80 &   .80 &  .81 \\
&     LSTM &  .74 &  .75 &  .77 &  .76 &  .77 &  .77 &  .78 &  .76 \\
&      GRU &  .76 &  .75 &  .77 &  .77 &  .76 &  .79 &  .77 &  .77 \\
&  MLP &  .73 &  .78 &  .76 &  .76 &  .78 &  .77 &  .76 &  .76 \\
&      CNN &  .74 &  .73 &  .77 &  .76 &  .77 &  .77 &  .78 &  .78 \\ \hline
\multirow{5}{*}{IMDB}            
&     BERT &  .92 &  .92 & .93 &  .92 &   .92 &  .92 &   .92 &  .92 \\
&     LSTM &   .90 &  .89 &  .89 &  .89 &  .89 &  .89 &  .89 &  .89 \\
&      GRU &   .90 &   .90 &  .89 &   .90 &  .89 &   .90 &  .89 &  .89 \\
&  MLP &  .88 &  .88 &  .88 &  .88 &  .89 &  .88 &  .89 &  .88 \\
&      CNN &  .89 &  .89 &   .90 &  .89 &  .89 &  .89 &  .89 &  .89 \\ \hline
\multirow{5}{*}{AG}            
&     BERT &  .95 &  .95 &  .94 &  .94 &  .95 &  .95 &  .94 &  .95 \\
&     LSTM &  .93 &  .93 &  .92 &  .93 &  .93 &  .93 &  .93 &  .93 \\
&      GRU &  .93 &  .93 &  .93 &  .93 &  .93 &  .93 &  .93 &  .93 \\
&  MLP &  .93 &  .93 &  .93 &  .92 &  .93 &  .92 &  .93 &  .93 \\
&      CNN &  .93 &  .93 &  .93 &  .93 &  .93 &  .93 &  .93 &  .93 \\\hline
\multirow{5}{*}{MIMIC}           

&     BERT &  .84 &  .83 &  .85 &  .84 &  .86 &  .84 &   .85 &  .83 \\                
&     LSTM &  .88 &  .89 &  .89 &  .89 &  .89 &   .90&   .90&   .90\\
&      GRU &  .89 &   .90&  .89 &  .89 &   .90&   .90&   .90&   .90\\
&  MLP &   .90&  .89 &  .88 &  .88 &  .89 &  .88 &  .89 &  .89 \\
&      CNN &   .90&  .89 &   .90&   .90&   .90&   .90&  .89 &   .90\\
\end{tabular}
\caption{Validation set F1-macro average scores (3 runs) across datasets, encoders and attention mechanisms for models with and without TaSc (No-TaSc). Standard deviations do not exceed 0.01.}
\label{tab:data_performances_validation}
\end{table}

\clearpage
\newpage
\onecolumn

\onecolumn
\clearpage
\newpage

\section{Comparing TaSc with Non-attention Input Importance Metrics \label{appendix:comparing_tasc}}

\begin{table*}[!h]
    \small
    \centering
    \setlength\tabcolsep{3pt}
\begin{tabular}{cc || ccc | cccc||ccc | cccc}
                      &         & \multicolumn{7}{c||}{\textbf{Tanh}}                           & \multicolumn{7}{c}{\textbf{Dot}}                                     \\
                      &         & \multicolumn{3}{c|}{\textbf{Non - TaSc}} & \multicolumn{4}{c||}{\textbf{Lin-TaSc}}              & \multicolumn{3}{c|}{\textbf{Non-TaSc}}            & \multicolumn{4}{c}{\textbf{Lin-TaSc}}             \\
\textbf{ Data} & \textbf{Enc()} & \textbf{WO}  & $\mathbf{x}\nabla \mathbf{x}$  &     \textbf{IG}  & \textbf{WO}  &     $\mathbf{x}\nabla \mathbf{x}$  &                \textbf{IG}  & $\alpha \nabla \alpha$  & \textbf{WO}  & $\mathbf{x}\nabla \mathbf{x}$  &       \textbf{IG}  &  \textbf{WO}  &      $\mathbf{x}\nabla \mathbf{x}$  &                 \textbf{IG}  & $\alpha \nabla \alpha$  \\ \hline

 \multirow{5}{*}{SST}
      &    BERT &            .29 &        .64 &          .51 &              .37 &   \underline{.30} &           \underline{.25} &           \textbf{.22} &  \textbf{.32} &       .62 &           .49 &              .35 &  \underline{.57} &                       .51 &                   .55 \\
      &    LSTM &            .25 &        .24 &          .20 &              .26 &              .33 &  \underline{\textbf{.19}} &                    \textbf{.19} &           .21 &       .23 &           \textbf{.19} &              .21 &  \textbf{\underline{.19}} &  \underline{\textbf{.19}} &                  \textbf{ .19} \\
      &     GRU &            .24 &        .22 &          .19 &              .29 &              .24 &                       .20 &           \textbf{.18} &           .24 &       .25 &           .23 &  \underline{.21} &  \textbf{\underline{.19}} &  \underline{\textbf{.19}} &                   \textbf{.19} \\
      &     MLP &            .36 &        .26 &          .24 &  \underline{.26} &   \underline{.20} &           \underline{.19} &           \textbf{.18} &           .22 &       .19 &           \textbf{.18} &              .24 &              .19 &                       .19 &          \textbf{.18} \\
      &     CNN &            .30 &        .25 &          .20 &  \underline{.27} &  \underline{.22} &            \underline{.20} &           \textbf{.19} &           .22 &       .20 &  \textbf{.18} &  \underline{.21} &   \underline{.20} &                       .20 &                   .19 \\ \hline

 \multirow{5}{*}{ADR}
 &    BERT &            .83 &        .91 &          .89 &  \underline{.73} &  \underline{.55} &           \underline{.38} &           \textbf{.31} &           .81 &       .90 &           .87 &  \underline{.68} &  \underline{.58} &           \underline{.52} &           \textbf{.50} \\
  &    LSTM &            .82 &        .81 &          .80 &  \underline{.54} &  \underline{.42} &           \underline{.34} &           \textbf{.32} &           .87 &       .88 &           .87 &  \underline{.42} &  \underline{.35} &  \underline{\textbf{.34}} &                   \textbf{.34} \\
  &     GRU &            .84 &        .84 &          .84 &  \underline{.49} &  \underline{.38} &           \underline{.36} &           \textbf{.35} &           .79 &       .80 &           .80 &   \underline{.50} &   \underline{.40} &           \underline{.44} &          \textbf{.38} \\
  &     MLP &            .71 &        .63 &          .57 &   \underline{.60} &  \underline{.36} &           \underline{.43} &           \textbf{.31} &           .49 &       .43 &  \textbf{.39} &  \underline{.49} &   \underline{.40} &                       .44 &                   .40 \\
  &     CNN &            .80 &        .78 &          .78 &  \underline{.57} &  \underline{.46} &           \underline{.39} &           \textbf{.37} &           .77 &       .74 &           .74 &  \underline{.52} &  \underline{.43} &           \underline{.38} &          \textbf{.36} \\ \hline
  
  \multirow{5}{*}{IMDB}
  &   BERT &            .23 &        .69 &      .43 &              .27 &  \underline{.14} &  \underline{.16} &           \textbf{.07} &           .24 &       .72 &     .49 &              .26 &  \underline{.27} &  \underline{\textbf{.17}} &                   \underline{.20} \\
  
  &    LSTM &            .18 &        .12 &      .07 &  \underline{.11} &              .13 &  \underline{.05} &           \textbf{.04} &           .26 &       .09 &     .07 &  \underline{.07} &  \underline{.06} &           \underline{.06} &          \textbf{.05} \\
    &     GRU &            .18 &        .12 &      .07 &  \underline{.11} &  \underline{.06} &  \underline{.05} &           \textbf{.04} &           .27 &       .15 &     .08 &  \underline{.09} &  \underline{\textbf{.05}} &  \underline{\textbf{.05}} &                   \textbf{.05} \\
    &     MLP &            .16 &        \textbf{.05} &      \textbf{.05} &  \underline{.07} &              \textbf{.05} &     \textbf{.05} &                    \textbf{.05} &           .18 &       .07 &     .06 &  \underline{.09} &  \underline{\textbf{.05}} &  \underline{\textbf{.05}} &                   \textbf{.05} \\
    &     CNN &            .21 &        .09 &      .07 &  \underline{.18} &  \underline{.07} &  \underline{.06} &           \textbf{.05} &           .27 &       .07 &     .06 &  \underline{.14} &  \underline{.07} &           \underline{.06} &          \textbf{.05} \\ \hline

  \multirow{5}{*}{AG}
  &    BERT &            .62 &        .78 &          .56 &              .64 &  \underline{.58} &           \underline{.54} &           \textbf{.50} &           .56 &       .76 &           .60 &              .56 &  \underline{.59} &  \underline{\textbf{.55}} &                   \underline{.60} \\
  &    LSTM &            .53 &        .51 &  \textbf{.30} &  \underline{.47} &  \underline{.37} &                       .31 &                    .38 &           .47 &       .52 &  \textbf{.35} &  \underline{.43} &   \underline{.40} &                       .36 &                   .46 \\
  &     GRU &            .45 &        .36 &          .31 &              .50 &   \underline{.30} &           \underline{.24} &            \textbf{.20} &           .54 &       .40 &           .30 &  \underline{.36} &  \underline{.24} &           \underline{.23} &          \textbf{.22} \\
  &     MLP &            .53 &        .24 &          .25 &              .53 &  \underline{.23} &           \underline{.23} &           \textbf{.19} &           .44 &       .25 &           .23 &   \underline{.40} &   \underline{\textbf{.19}} &                       .25 &           \textbf{.19} \\
  &     CNN &            .55 &        .38 &          .28 &  \underline{.48} &  \underline{.29} &           \underline{.24} &            \textbf{.20} &           .53 &       .35 &           .25 &  \underline{.39} &  \underline{.27} &           \underline{.23} &          \textbf{.21} \\ \hline
  
  \multirow{5}{*}{MIMIC}
  &     BERT &            .24 &        .67 &      .43 &              .31 &   \underline{.10} &  \underline{.04} &           \textbf{.03} &           .21 &       .57 &     .26 &             .25 &  \underline{.07} &  \underline{\textbf{.05}} &          \textbf{.05} \\
  
  &    LSTM  &            .35 &        .32 &      .12 &              .37 &  \underline{\textbf{.01}} &           \underline{.02} &                    \textbf{.01}   &     .28 &       .40 &     .30 &             .40 &  \underline{\textbf{.01}} &  \underline{.02} &          \textbf{.01} \\
  
  &  GRU &            .20 &        .24 &      .23 &              .46 &  \underline{\textbf{.01}} &  \underline{.02} &           \textbf{.01} &           .36 &       .18 &     .08 &             .42 &  \underline{\textbf{.01}} &  \underline{.02} &          \textbf{.01} \\
  
  &     MLP &            .40 &        .03 &      .22 &  \underline{.18} &  \underline{\textbf{.01}} &           \underline{.02} &                    \textbf{.01} &           .13 &       .04 &     .03 &             .16 &  \underline{\textbf{.02}} &  \underline{\textbf{.02}} &                   \textbf{.02} \\
  &     CNN &            .26 &        .15 &      .02 &              .52 &           \textbf{\underline{.01}} &  \underline{\textbf{.01}} &                    \textbf{.01} &           .43 &       .09 &     \textbf{.02} &             .49 &  \underline{.03} &  \underline{\textbf{.02}} &                   \textbf{.02} \\ \\
  
\end{tabular}
\caption{Average fraction of tokens required to cause a decision flip using the best performing attention-based ranking ($\alpha \nabla \alpha$) with TaSc, \textit{Word omission}, (\textbf{WO}), \textit{InputXGrad}, ($\nabla \mathbf{x}$) and \textit{Integrated Gradients} ($\mathbf{IG}$). \underline{Underlined} values denote that Lin-TaSc is better and \textbf{bold} values denote the best performing method row-wise. (lower is better)}
\label{tab:tasc_gradient_comparison_full}
\end{table*}

\end{appendices}

\end{document}